\documentclass[nohyperref]{article}

\usepackage{microtype}
\usepackage{graphicx}
\usepackage{subfigure}
\usepackage{booktabs} 

\usepackage{hyperref}

\usepackage[accepted]{icml2022}

\usepackage{amsmath}
\usepackage{amssymb}
\usepackage{mathtools}
\usepackage{amsthm}

\usepackage[capitalize,noabbrev]{cleveref}

\theoremstyle{plain}

\theoremstyle{definition}

\theoremstyle{remark}

\usepackage[textsize=tiny]{todonotes}

\icmltitlerunning{Interpretable Distribution Shift Detection using Optimal Transport}

\begin{document}

\twocolumn[
\icmltitle{Interpretable Distribution Shift Detection using Optimal Transport}



\icmlsetsymbol{equal}{*}

\begin{icmlauthorlist}
\icmlauthor{Neha Hulkund}{yyy}
\icmlauthor{Nicolo Fusi}{comp}
\icmlauthor{Jennifer Wortman Vaughan}{comp}
\icmlauthor{David Alvarez-Melis}{comp}
\end{icmlauthorlist}

\icmlaffiliation{yyy}{Massachusetts Institute of Technology. Work completed while author was an intern at Microsoft Research}
\icmlaffiliation{comp}{Microsoft Research}

\icmlcorrespondingauthor{Neha Hulkund}{nhulkund@mit.edu}

\icmlkeywords{Machine Learning, ICML}

\vskip 0.3in
]



\printAffiliationsAndNotice{}  

\begin{abstract}
We propose a method to identify and characterize distribution shifts in classification datasets based on optimal transport. It allows the user to identify the extent to which each class is affected by the shift, and retrieves corresponding pairs of samples to provide insights on its nature. We illustrate its use on synthetic and natural shift examples. While the results we present are preliminary, we hope that this inspires future work on interpretable methods for analyzing distribution shifts.
\end{abstract}
\vspace{-2mm}
\section{Introduction and Related Work}
As machine learning models become more widely used in high-stakes domains such as healthcare and education, it is important to analyze situations when models fail. This often occurs with \textit{distribution shift}, e.g.,  when train and test distributions differ. For example, a model could be trained on MRI data from one machine and tested on data from another, leading to performance drops that negatively affect patient care and derail trust. Detecting and identifying these distribution shifts are important steps in deciding whether to apply mitigation techniques, such as domain adaption or stronger regularization. While out-of-distribution detection techniques identify which samples are different than the training distribution, they give little reason as to \textit{why} the shifts might be occurring \cite{OOD_ODIN, EnergyOOD, RobustOOD}. 
Much previous research has been focused on interpreting model decisions from analyses on model gradients \cite{GradCam, Saliency1, Saliency2} to post-hoc model explanations like feature summary statistics \cite{Fisher2018ModelCR, ShapleyValue}. 
However, there has been comparatively less work in interpretability from the dataset point of view, especially under the conditions of dataset distribution shift. In this work, we propose a method to identify and characterize distribution shift through examples using optimal transport (OT) \cite{peyre2019computational}. 

\section{Methodology}
Consider two datasets  $\{\mathbf{x}_a^{i}\}_{i=1}^n$ and $\{\mathbf{x}_b^{j}\}_{j=1}^m$ with associated probability weights $\{\mathbf{a}\in \Delta_n, \mathbf{b}  \in \Delta_m\}$. When interpreted as empirical distributions $\alpha_n,\beta_m$, OT  provides a way to compare them by solving the problem 
$$\mathrm{OT}(\alpha_n, \beta_m)= \min_{\pi \in \Pi(\alpha,\beta)} \sum_{i,j=1}^{n,m} d(\mathbf{x}_a^{i}, \mathbf{x}_b^{j}) \pi_{ij},$$
where $\Pi(\alpha,\beta)$ is the set of \textit{transport couplings} (joint distributions) with $\mathbf{a}$ and $\mathbf{b}$ as marginals. The minimizing $\pi^*$ can be interpreted as the `best' (least-cost) soft matching between the two samples, and its total cost, i.e., $\mathrm{OT}(\alpha_n, \beta_m)$, yields a notion of distance between them. Although typically formulated between unlabeled samples, a recent extension of OT allows for comparison of classification datasets, called the optimal transport dataset distance (OTDD) \cite{AlvarezMelis2020GeometricDD}.




Here, we propose to use OTDD to provide examples of distribution shift. We first compute OTDD between the original and shifted datasets, and then use the optimal coupling $\pi^*$ to identify corresponding images across the two datasets, ranking them by their dissimilarity. Specifically, we select pairs $(i,j)$ for which $\pi^*_{ij}$ is high, and visualize them ranked according to the value of $ d(\mathbf{x}_a^{i}, \mathbf{x}_b^{j})\cdot \pi^*_{ij}$. Farthest and closest pairs indicate where there is most and least shift, respectively. Furthermore, the fact that these pairs are the `best corresponding' (in an OT sense) across the datasets controls for natural variation and highlights shift-induced change, potentially making them interpretable and providing insights on the nature of the shift.




\section{Synthetic MNIST example}
As a proof of concept, we use a synthetic shift of Gaussian noise applied to one class of the handwritten digits dataset MNIST.
 
 \begin{figure}[h]
\centering
\includegraphics[width=1.0\linewidth]{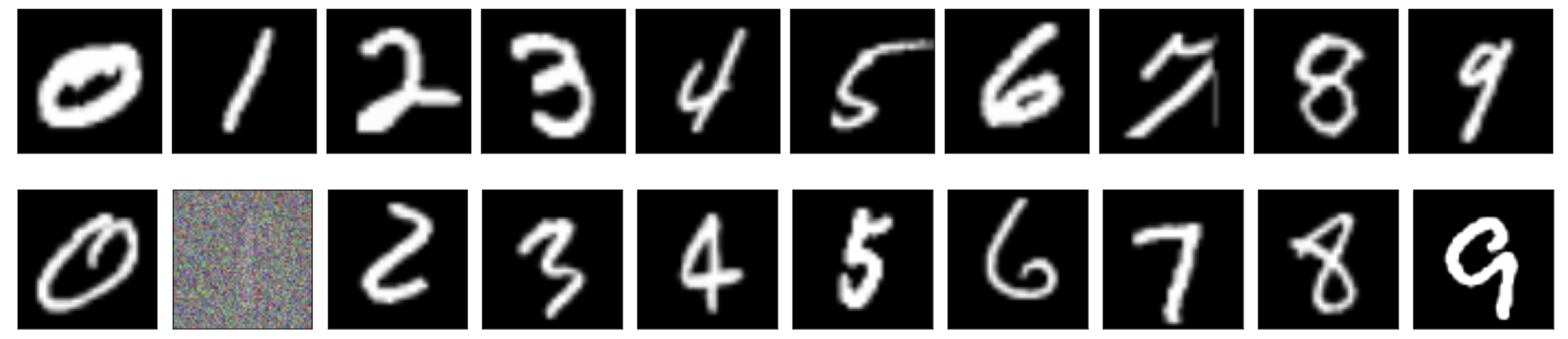}
\caption{MNIST shift dataset, where the top row is the classic training dataset and the second row is the shifted testing dataset with Gaussian noise added to class 1.} \label{fig:MNIST_dataset}
\end{figure}
 
\begin{figure}
\centering
\includegraphics[width=0.49\linewidth]{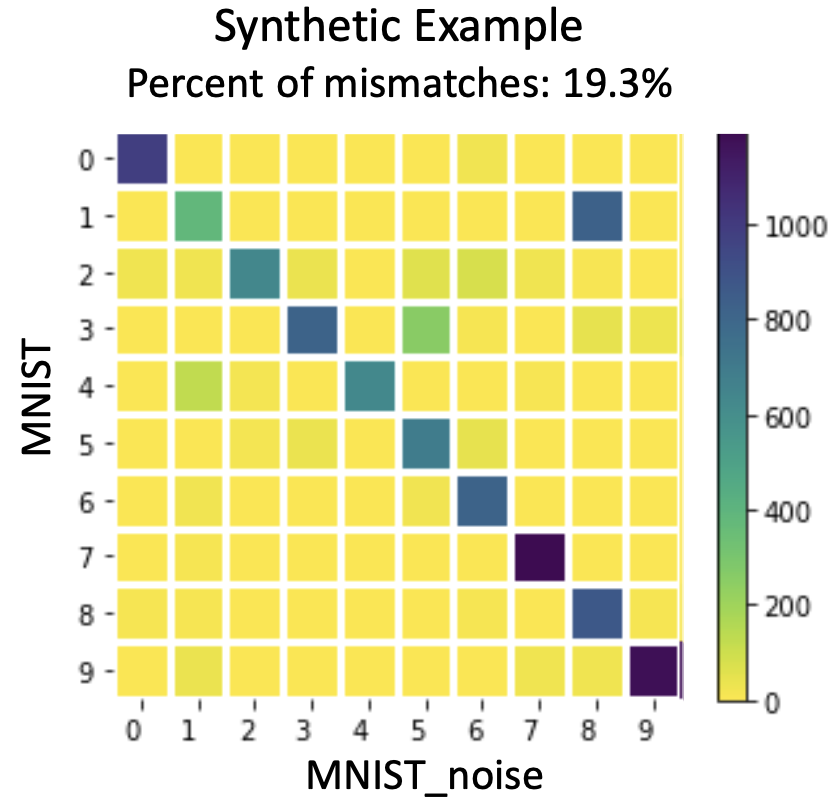}%
\includegraphics[width=0.49\linewidth]{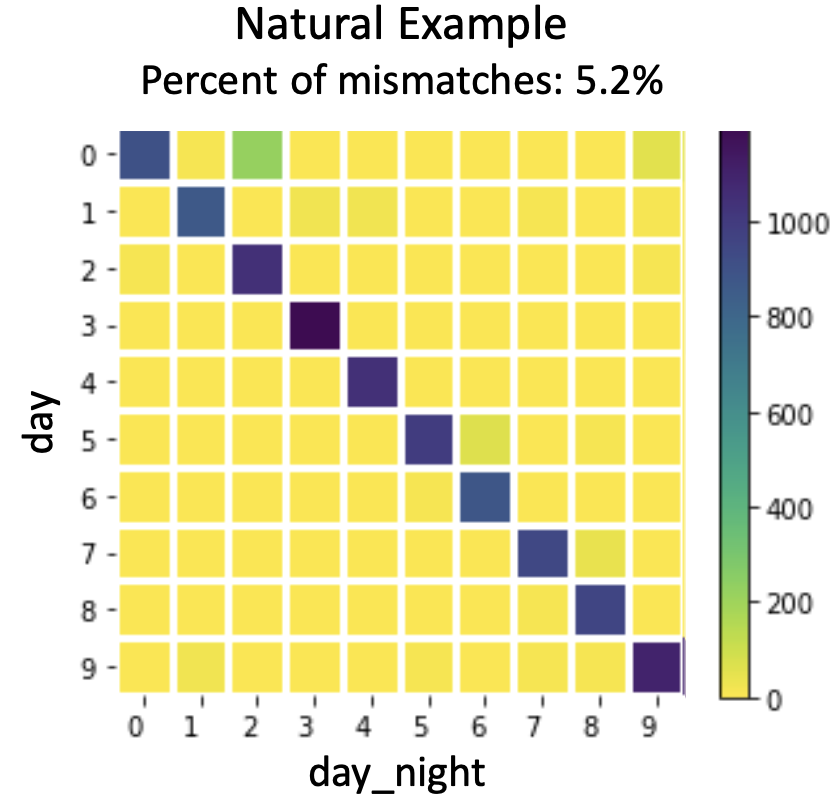}
\vspace{-0.35cm}
\caption{Coupling matches for synthetic and natural examples.}
\label{fig:heatmap}
\end{figure}

\begin{figure}[h]
\centering
\includegraphics[width=0.9\linewidth, ]{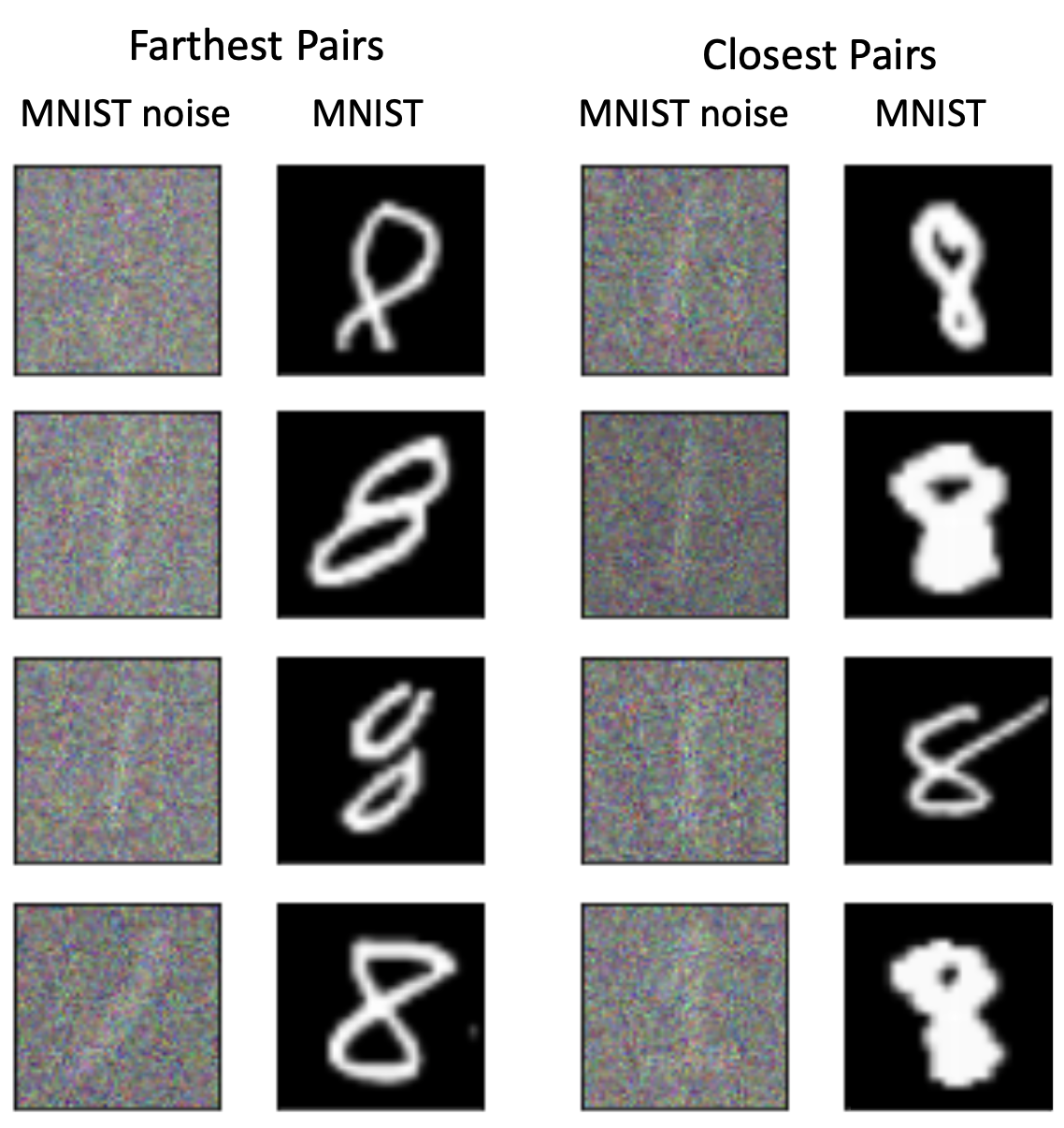}
\caption{MNIST pairs between class 1 and 8.} 
\label{fig:MNIST_pairs}
\end{figure}

 From the heatmap of matches in our coupling matrix $\pi^*$ in \cref{fig:heatmap}, we find that most classes match well, with the exception of class 1 from the noisy dataset to class 8 from the original dataset. This indicates that a noisy 1 looks more similar to an noiseless 8 than an noiseless 1, representing a type of dataset shift. We can now examine this class mismatch more closely, taking the OT coupling between the two classes to obtain the closest and farthest ranked pairs \cref{fig:MNIST_pairs}. From the pairings, the closest pairs of noiseless images from class 8 have less definitive shape to them, possibly creating mismatch between the classes. Furthermore, we found our mismatch finding to be consistent with classifier prediction, where noiseless 8's are often mislabeled as 1's under this setting. We believe that further analyses on shape similarities on image datasets between classes can lead to insight about class-level shifts.

\section{Day/Night Natural example}
\begin{figure}[h]
\centering
\includegraphics[width=1.0\linewidth, trim = {0 0.9cm 0 0.3cm}]{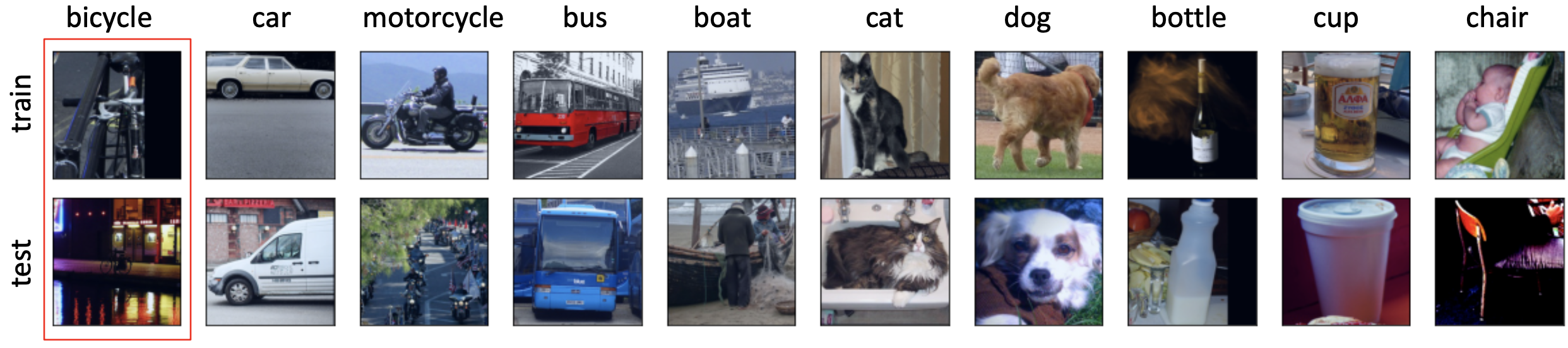}
\caption{CODaN dataset shift, with class 0 (bicycle) images  taken during the day in the training set and at night in the testing set.} 
\label{fig:CODAN_dataset}
\end{figure}

\begin{figure}[h]
\centering
\includegraphics[width=1.0\linewidth, trim = {0 0.9cm 0 0.3cm}]{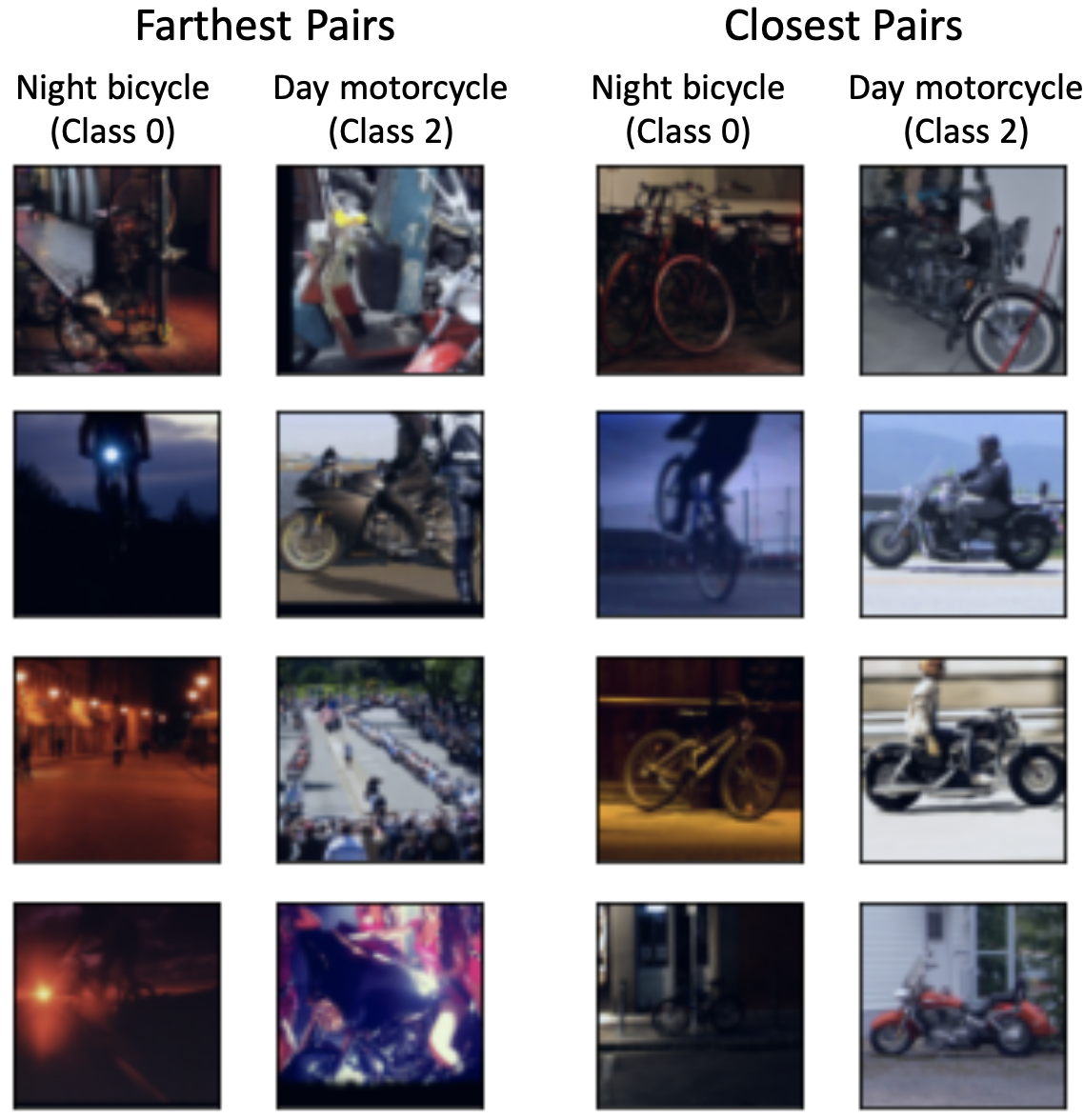}
\caption{CODaN pairs between night-bicycle and day-motorcyle.} 
\label{fig:CODAN_pairs}
\end{figure}

As a natural shift, we use the Common Objects Day and Night (CODaN) dataset, an image classification dataset of 10 common object classes recorded in both day and nighttime \cite{CODAN}. The training dataset is the 10 classes taken from the day distribution and the testing dataset is 9 classes taken from the day distribution and 1 class (bicycle) from the night distribution, as seen in \cref{fig:CODAN_dataset}. In the heatmap of matched classes in \cref{fig:heatmap}, we can see a mismatched class pair $\{0,2\}$, or the night bicycle/day motorcycle pair. This indicates that a bicycle at night is matched more closely to a motorcycle in the day than it is to itself in the day. The coupling pairs in \cref{fig:CODAN_pairs} illustrate which bicycle-motorcycle pairs are closest and farthest in the dataset, implying that at night with lower image brightness, bicycles look darker, similar to motorcycles during the daytime.

\vspace{0.00mm} 
\vspace{0.00mm}

\section{Discussion}
The method we propose characterizes distribution shifts through OT distances and image correspondences across datasets. Since these examples are best corresponding, there is potential that they might offer qualitative interpretability about dataset shift. In a domain such as healthcare, examples of such pairings could then be provided to domain experts to gain insights of possible causes of the data shift when model explanations are not sufficient. While other methods illustrate examples of data that occur on an example level, this method could be potentially useful for detecting class level drifts within a dataset. Further work remains to evaluate how this method works on other datasets and whether it does indeed lead to greater insight. 

\bibliography{example_paper}
\bibliographystyle{icml2022}

\end{document}